\documentclass{article}

 \usepackage[preprint]{neurips_2026}

\usepackage[utf8]{inputenc} 
\usepackage[T1]{fontenc}    
\usepackage{hyperref}       
\usepackage{url}            
\usepackage{booktabs}       
\usepackage{amsfonts}       
\usepackage{nicefrac}       
\usepackage{microtype}      
\usepackage{xcolor}         

\usepackage{enumitem}
\usepackage{algorithm}
\usepackage{algorithmic}
\usepackage{amsmath}
\usepackage{amssymb}
\usepackage{mathtools}
\usepackage{amsthm}
\usepackage{comment}
\usepackage{multirow}
\usepackage{stfloats}
\usepackage{wrapfig}

\title{
Unicorn: Scaling High-Dimensional Time Series Forecasting via Universal Correlation Modeling
}

\author{
\textbf{Haochen Yuan} \quad
\textbf{Yichen Song} \quad
\textbf{Yunbo Wang}\thanks{Corresponding author.} \quad
\textbf{Xiaokang Yang}\\[0.5em]
MoE Key Lab of Artificial Intelligence, AI Institute, School of Computer Science\\
Shanghai Jiao Tong University\\
\texttt{\{yuanhaochen, syc.x\_x, yunbow, xkyang\}@sjtu.edu.cn}
}

\begin{document}

\maketitle

\begin{abstract}
Modern time series architectures face a fundamental trade-off: channel-independent models scale well with increasing data volume but ignore critical inter-channel dependencies, while channel-dependent models are expressive but remain ``dimension-bounded'', struggling to generalize across heterogeneous datasets.
To bridge this gap, we introduce \textbf{Unicorn} (Universal Correlation Network), a framework for scalable, multi-dataset pretraining on high-dimensional time series.
At the core of Unicorn is a \emph{latent prototype codebook} that decouples correlation modeling from specific channel identities. By projecting heterogeneous channels into a shared latent space, UniCorN learns \emph{identity-agnostic, reusable interaction patterns} that transfer across domains with diverse dimensionalities and semantics.
Extensive experiments show that Unicorn significantly outperforms state-of-the-art forecasting architectures, particularly in few-shot transfer scenarios, offering a scalable path toward multivariate time series foundation models.

\end{abstract}

\begin{wrapfigure}{r}{0.47\textwidth}
    \vspace{-10pt}
    \centering
    \includegraphics[
        width=\linewidth
    ]{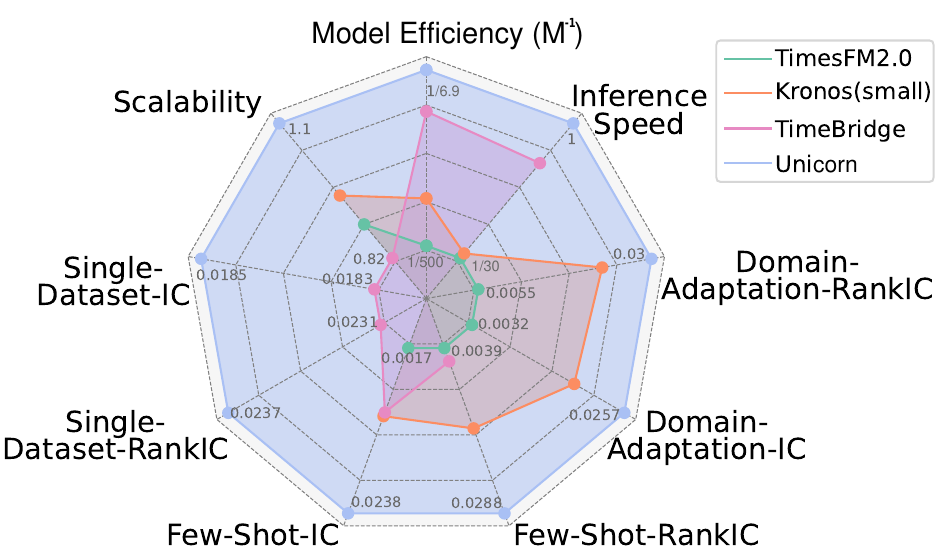}
    \vspace{-15pt}
    \caption{\textbf{Performance on a 587-stock A-share dataset.}
    The radar chart benchmarks model performance across four key dimensions: efficiency, generalization, scalability, and few-shot adaptability.
    Scalability is evaluated by the performance delta between full-channel and $25\%$ channel configurations during finetuning and inference.
    Few-shot performance is measured by finetuning on a restricted $25\%$ subset of the training data.}
    \vspace{-23pt}
\end{wrapfigure}

\section{Introduction}

High-dimensional time series forecasting is at the heart of many real-world applications, ranging from financial market analysis to large-scale energy system management and climate modeling.
In these domains, systems must jointly reason over hundreds or even thousands of correlated variables (\textit{i.e.}, input channels). 
Despite recent advances in time series foundation models, scaling modern architectures to such high-dimensional settings remains challenging.
This challenge arises from the fundamental conflicts between \textbf{domain scalability} and \textbf{correlation modeling}: On one hand, channel-independent (CI) models~\citep{kronos,timesfm} process each variable independently with shared parameters, which scales well across heterogeneous datasets but discards inter-channel interactions.
On the other hand, channel-dependent (CD) models~\citep{timebridge,fedformer} explicitly capture interactions among variables, typically through multi-head attention mechanisms that are inherently tied to specific channel identities and fixed dimensionalities.

Despite ignoring inter-channel dependencies, recent empirical studies show that CI models often outperform their CD counterparts in multivariate forecasting tasks~\citep{kronos}. 
This seemingly paradoxical result suggests a key limitation of existing CD approaches: by tightly coupling correlation modeling to fixed channel identities, they learn dataset-specific dependencies that fail to generalize.
This issue is particularly pronounced in high-dimensional regimes, where correlations are sparse and unevenly observed, making CD models prone to overfitting.
For example, in financial markets where the number of sequences is smaller than the number of assets ($M < N$), CD models struggle to distinguish structural signals from noise, while CI models benefit from an implicit data multiplication effect ($M \times N$), yielding stronger generalization.

These observations motivate a central hypothesis: the limitation of CD models lies \textbf{NOT} in modeling correlations per se, but in the \textbf{scale and manner} in which they are learned.
Building on this insight, we propose \textit{Unicorn} (Universal Correlation Network), a unified framework that decouples correlation modeling from explicit channel identities to enable scalable pretraining across heterogeneous datasets.
Our design starts with a basic assumption: while physical channel identities vary across domains, the underlying interaction laws (\textit{e.g.}, similar sector-level co-movements across different stock markets) often present universal properties.
Unicorn achieves this with a learnable \textbf{Latent Prototype Codebook}, which functions as a shared set of abstract anchors for the joint modeling of diverse variables. To align heterogeneous channels with these anchors, we introduce a \textbf{Spectral Global Guidance} module that employs frequency-informed features to bridge semantic gaps.
Rather than parameterizing direct \textit{channel-to-channel} interactions, UniCorN routes dependencies through a \textit{channel-to-prototype} interaction mechanism. By treating prototypes as an intermediate ``interaction bottleneck'', the model captures complex relational structures in an identity-agnostic manner, effectively bridging the gap between cross-domain scalability and expressive inter-channel correlation modeling.

%
%
%
%
%

Extensive experiments demonstrate that Unicorn significantly outperforms state-of-the-art CI foundation models, including TimesFM~\citep{timesfm} and Kronos~\citep{kronos}, as well as specialized CD architectures.
We empirically demonstrate that Unicorn effectively mitigates the overfitting issues of CD methods and shows strong scaling effects as the pretraining data volume increases.

\vspace{-4pt}
\section{Problem Formalization}
\vspace{-2pt}

We consider multi-dataset pretraining for high-dimensional time series forecasting.
Given an input sequence $\mathbf{X} \in \mathbb{R}^{C \times I}$ spanning a lookback window of length $I$ across $C$ variables (channels), our objective is to predict the future values $\mathbf{Y} \in \mathbb{R}^{C \times O}$ for a horizon of length $O$.
This high-dimensional setting poses two fundamental challenges: 
First, standard CD models incur $O(C^2)$ complexity. When $C$ is large (\textit{e.g.}, hundreds or thousands) and exceeds temporal observations ($C > I$), these models suffer from parameter explosion and severe overfitting. 
Second, in large-scale pretraining, $C$ varies across datasets ($C \in \{C_1, C_2, \dots\}$). Fixed-dimension interaction matrices ($\mathbb{R}^{C \times C}$) are incompatible with joint training on heterogeneous channel sets.


\vspace{-4pt}
\section{Related Work}
\label{sec:related_work}
\vspace{-2pt}


Multivariate time series forecasting is commonly organized around \emph{channel independent} (CI) and \emph{channel dependent} (CD) paradigms. 
CI methods model each variate separately, which often improves robustness and scalability under heterogeneous short-term dynamics~\citep{rangapuram2018deep, park2022learningquantilefunctionsquantile, sagheer2019time, li2019enhancing, zhou2021informer}. 
Representative CI models include TiDE~\citep{tide}, Non-stationary Transformers~\citep{liu2022non}, TEMPO~\citep{cao2023tempo}, PDF~\citep{pdf}, and SparseTSF~\citep{sparsetsf}. 
While CD methods explicitly exploit inter-channel dependencies, direct coupling often degrades generalization~\citep{wang2019deep, wang2022micn, zhou2023one, yi2024frequency, wang2023timemixer}.
Existing CD research can be broadly categorized into three primary directions. First, sequential and probabilistic models, such as DeepAR \citep{flunkert2017deepar}, GRU-D \citep{che2018recurrent}, and TimeGrad \citep{rasul2021autoregressive}, capture correlations by sharing global temporal dynamics.
Second, graph-based methods, such as MTGNN~\citep{wu2020connectingdotsmultivariatetime}, explicitly model relational structures to facilitate information propagation across channels.
Finally, attention- or mixing-based architectures directly aggregate variate-level representations, including iTransformer~\citep{itransformer}, Crossformer~\citep{crossformer}, TimeMixer~\citep{wang2024timexer}, and a range of related approaches~\citep{zhou2024scalable, patchtst, autoformer, fedformer, timesnet}.
More recent methods, such as TimeBridge~\citep{timebridge}, further incorporate cointegration-aware attention to model long-term dependencies, while SOFTS~\citep{softs} introduces an efficient aggregate-redistribute mechanism for scalable cross-variate modeling.

Time series foundation models aim at strong zero-shot and few-shot performance across domains with large-scale pretraining on massive time series data.
Typical architectures include decoder-only~\citep{timesfm, zhou2023one, liu2024timer, xiaoming2025time, liu2025sundial, ansari2024chronos}, encoder-only~\citep{goswami2024moment, woo2024unified}, and encoder-decoder models~\citep{garza2023timegpt}.
%
%
TimesFM~\citep{timesfm} and Kronos~\citep{kronos} are designed with CI architectures that process multivariate inputs without explicitly modeling inter-channel dependencies. While they can handle datasets with varying channel dimensions, they do not differentiate between channels during pretraining, focusing solely on temporal dependencies.
Unlike TimesFM and Kronos, our method overcomes this limitation through domain-invariant prototype interaction, enabling channel-dependent modeling and multi-dataset pretraining simultaneously, which makes it particularly well-suited for high-dimensional forecasting.


\vspace{-4pt}
\section{Method}
\vspace{-2pt}


To enable domain scalability, Unicorn is founded upon a guiding principle: 
\textit{inter-channel dependencies should be modeled in a shared latent interaction space rather than via explicit channel-to-channel parameterization.}
The core idea is to project heterogeneous high-dimensional channels into a compact set of $K$ learnable prototypes ($K \ll C$), forming a universal correlation vocabulary that scales to arbitrary dimensions and generalizes across diverse domains.

Concretely, Unicorn consists of three components: 
(i) a channel-wise encoder that extracts univariate temporal representations (Section \ref{sec:stage1});
(ii) a frequency-informed global feature module that injects spectral cues to guide representation learning (Section \ref{sec:stage2}); and
(iii) a latent interaction module that models channel-to-prototype dependencies in the latent space (Section \ref{sec:stage3}).
We further detail the training pipeline in Section \ref{sec:train} and analyze computational complexity in Section \ref{sec:complexity}.

\begin{figure*}[t]
	\centering
	\includegraphics[width=0.99\linewidth]{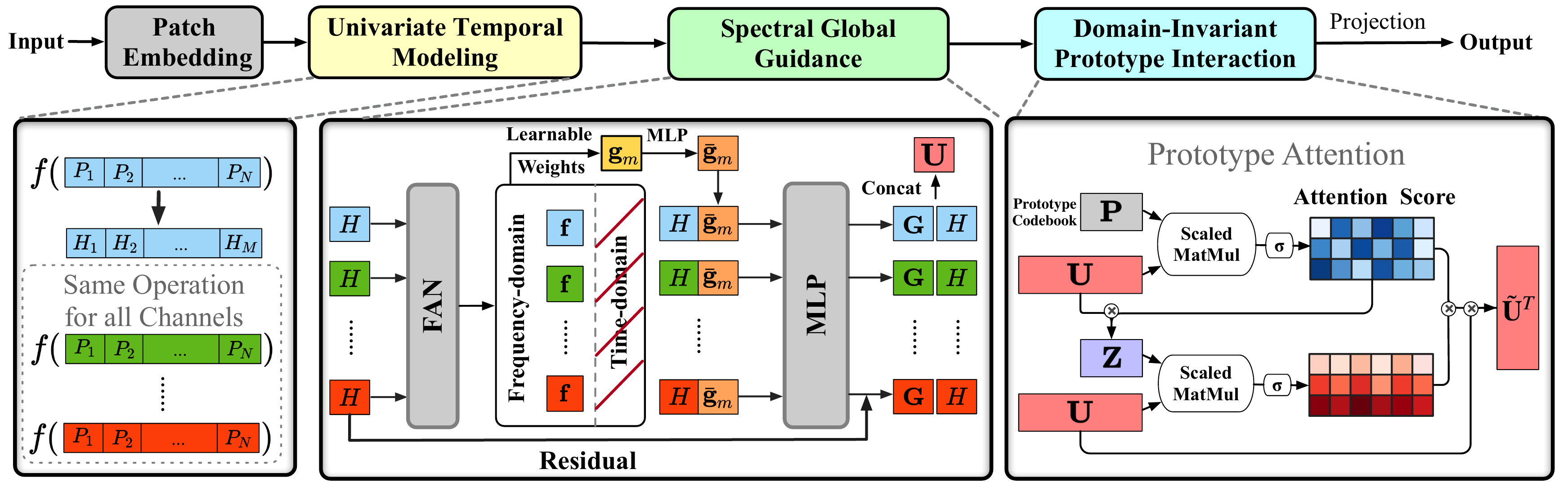}
    \vspace{-5pt}
	\caption{\textbf{The Unicorn architecture.} 
    \textit{Temporal Extraction (Left):} 
    Scalable univariate modeling via shared patch embedding.
    \textit{Frequency Guidance (Middle):} A Fourier analysis network generates spectral features ($\bar{g}_m$) to guide channel alignment. 
    \textit{Prototype Interaction (Right):} Channel-to-prototype cross-attention ($U$, $P$) yields identity-decoupled inter-channel features ($\tilde{U}$).
    }
	\label{fig:method}
    \vspace{-5pt}
\end{figure*}

\vspace{-4pt}
\subsection{Temporal Modeling}
\label{sec:stage1}
\vspace{-2pt}
Given $\mathbf{X}\in\mathbb{R}^{C\times I}$, we partition each channel into non-overlapping patches of length $S$, resulting in $N=\lceil I/S\rceil$ patches. Each temporal patch is linearly projected into a $D$-dimensional token, producing patch tokens $\mathbf{P}\in\mathbb{R}^{C\times N\times D}$.
Unicorn adopts TimeBridge~\citep{timebridge} as the temporal backbone for channel-independent modeling, and applies Integrated Attention to patches \textit{within each channel} to capture intra-channel temporal dependencies, while mitigating short-term non-stationarity.

To model long-term dependencies, patch tokens are further downsampled along the temporal dimension from $N$ to $M$ tokens ($M\ll N$), producing long-term tokens $\mathbf{H} \in \mathbb{R}^{C \times M \times D}$ with $M \ll I.$
Each slice $\mathbf{H}_{:,m}$ corresponds to a coarse temporal abstraction at index $m$ and serves as the input to subsequent inter-channel modules.
Up to this stage, all operations are channel-independent, which supports pretraining across datasets with heterogeneous channel sets.

\vspace{-4pt}
\subsection{Spectral Global Guidance}
\label{sec:stage2}
\vspace{-2pt}

Domain-reusable channel modeling requires stable interaction cues, typically manifesting as recurring periodic patterns across domains. We address this with a frequency-guided module that leverages spectral information to construct a domain-transferable inter-channel context.

\vspace{-8pt}
\paragraph{Spectral feature extraction.}
To explicitly capture periodic structures, we employ a Fourier Analysis Network (FAN)~\citep{fan} as a shared frequency extractor.
Given $\mathbf{H} \in \mathbb{R}^{C \times M \times D}$ from the channel-wise encoder, the FAN operates in the frequency domain and extracts periodicity-related representations by filtering out non-periodic temporal variations.
Specifically, for each channel and long-term index, the FAN produces a spectral descriptor $\mathbf{f}_{c,m}$ that encodes dominant periodic signatures:
\begin{equation}
\mathbf{f}_{c,m} = \mathrm{FAN}(\mathbf{H}_{c,m,:}) \in \mathbb{R}^{d_f},
\end{equation}
where the parameters of the FAN are shared across all channels and datasets to ensure a universal feature space. Unlike traditional approaches that mix channels in the time domain, our design recognizes that local temporal dynamics are already sufficiently captured by the shared backbone. By performing interaction in the frequency domain, we focus on stable, global characteristics that serve as reliable anchors for cross-dataset alignment.

\vspace{-8pt}
\paragraph{Aggregate-redistribute mechanism.}
Building on these spectral descriptors, we synthesize a global inter-channel context with an aggregate-redistribute mechanism, without introducing dataset-specific model parameters. For each patch index $m$, we compute normalized channel importance scores $a_{c,m}$ based on their frequency characteristics:
\vspace{-2pt}
\begin{equation}
s_{c,m} = \mathbf{w}^{\top}\mathbf{f}_{c,m},\quad
a_{c,m} = \frac{\exp(s_{c,m})}{\sum_{j=1}^{C}\exp(s_{j,m})},
\end{equation}
where $\mathbf{w}\in\mathbb{R}^{d_f}$ is learnable and $\sum_{c=1}^{C} a_{c,m} = 1$ for each $m$.
These scores determine the relative contribution of each channel to a shared global representation at each index $m$:
\vspace{-2pt}
\begin{equation}
\mathbf{g}_m = \sum_{c=1}^{C} a_{c,m}\,\mathbf{H}_{c,m,:}\in\mathbb{R}^{D}.
\end{equation}
The aggregated global token $\mathbf{g}_m$ is refined via a dimension-wise MLP denoted by $\phi(\cdot)$, yielding the refined context $\bar{\mathbf{g}}_m = \phi(\mathbf{g}_m)\in\mathbb{R}^{D}$. This context is subsequently redistributed to individual channels through a residual fusion process:
\begin{equation}
\mathbf{G}_{c,m,:} = \mathbf{H}_{c,m,:} + \psi\big([\mathbf{H}_{c,m,:};\bar{\mathbf{g}}_m]\big)\in\mathbb{R}^{D},
\end{equation}
where $[\cdot\,;\cdot]$ denotes concatenation along the feature dimension, and $\psi(\cdot)$ is an MLP shared across channels.
This stage concludes by enriching the original CI features with spectral-aware global context. For each $m$, we construct the augmented representation:
\begin{equation}
\mathbf{U}_{:,m,:} = [\mathbf{H}_{:,m,:};\mathbf{G}_{:,m,:}] \in \mathbb{R}^{C\times 2D}.
\end{equation}
By preserving the univariate temporal features while augmenting them with domain-stable periodic cues, this concatenated representation $\mathbf{U}$ serves as the optimal input for the subsequent Prototype-Mediated Interaction, facilitating the alignment of heterogeneous channels.

\vspace{-4pt}
\subsection{Channel-Prototype Interaction}
\label{sec:stage3}
\vspace{-2pt}

%
To learn transferable inter-channel dependencies, Unicorn introduces a latent prototype codebook and models interactions within a domain-reusable latent space.
This architecture replaces identity-specific $C \times C$ interactions with (i) \textit{a set of latent channel interaction prototypes} and (ii) \textit{a dataset-adaptive channel-prototype aggregation and redistribution mechanism.}

\vspace{-8pt}
\paragraph{Latent prototype codebook.}
Given the spectral-temporal representations $\mathbf{U}_{:,m,:} \in \mathbb{R}^{C \times 2D}$ from the previous stage, we maintain a learnable set of latent prototypes $\mathbf{P} \in \mathbb{R}^{K \times 2D}$, where $K$ is fixed and independent of the channel dimension ($K \ll C$). 
We treat $\mathbf{P}$ as a latent codebook where each prototype represents an abstract interaction anchor shared across datasets, while channels are softly assigned to these anchors based on dynamic content.
For each patch index $m$, the inter-channel interaction is computed via a cascaded attention mechanism, which follows a structured aggregate-redistribute pipeline.

\vspace{-8pt}
\paragraph{Channel-to-prototype aggregation.}
The prototypes act as queries to \textit{read} from channel-wise representations, forming prototype-specific summaries:
\vspace{-4pt}
\begin{equation}
\begin{split}
    \mathbf{Z}_m &= \mathrm{Attn}(\mathbf{P}, \mathbf{U}_{:,m,:}, \mathbf{U}_{:,m,:}) \in \mathbb{R}^{K \times 2D} =\mathrm{softmax}\!\left(\frac{\mathbf{P}\mathbf{U}_{:,m,:}^{\top}}{\sqrt{2D}}\right) \cdot \mathbf{U}_{:,m,:}
\end{split}
\label{eqn:protype0}
\end{equation}
The content-driven assignment matrix $\mathbf{A}_m=\mathrm{softmax}({\mathbf{P}\mathbf{U}_{:,m,:}^{\top}}/{\sqrt{2D}})\in\mathbb{R}^{K\times C}$ groups channels with similar temporal and spectral patterns under shared prototypes, producing $K$ latent interaction tokens that are shared across datasets.

\vspace{-8pt}
\paragraph{Prototype-to-channel redistribution.}
Once the channel dependencies are computed within the compact prototype space, this stage redistributes the refined signals back to the original channels.
Specifically, we reuse spectral-temporal augmented representations to query the prototype summaries and retrieve interaction-enhanced features:
\begin{equation}
\begin{split}
    \tilde{\mathbf{U}}_{:,m,:} &= \mathrm{Attn}(\mathbf{U}_{:,m,:}, \mathbf{Z}_m, \mathbf{Z}_m) \in \mathbb{R}^{C \times 2D}  =\mathrm{softmax}\!\left(\frac{\mathbf{U}_{:,m,:}\mathbf{Z}_m^{\top}}{\sqrt{2D}}\right) \cdot \mathbf{Z}_m.
\end{split}
\label{eqn:protype1}
\end{equation}
The channel-to-prototype redistribution matrix $\mathbf{B}_m=\mathrm{softmax}({\mathbf{U}_{:,m,:}\mathbf{Z}_m^{\top}}/{\sqrt{2D}})\in\mathbb{R}^{C\times K}$  propagate the mediated interactions back to the original channel space.
%
%
The resulting interactive features are integrated with the original representations via residual connections and a dimension-wise MLP. This creates a refined, spectral-temporal augmented feature set.

\vspace{-4pt}
\subsection{Training Scheme}
\label{sec:train}
\vspace{-2pt}

\paragraph{Spectral-temporal objective.}
To ensure the model captures both point-wise temporal fidelity and global periodic structures, Unicorn is supervised through a dual-domain objective.
Given the ground-truth sequence $\mathbf{Y}\in\mathbb{R}^{C\times O}$ and the prediction $\hat{\mathbf{Y}}$, we define the forecasting loss as:
\vspace{-4pt}
\begin{equation}
\mathcal{L}_{\text{pred}}
= \  (1-\alpha)\frac{1}{B}\sum_{i=1}^{B}\left|Y_i-\hat{Y}_i\right| 
  + \alpha\frac{1}{B}\sum_{i=1}^{B}\left|\mathrm{FFT}(Y_i)-\mathrm{FFT}(\hat{Y}_i)\right|,
\label{eq:loss}
\end{equation}
where $\mathrm{FFT}(\cdot)$ denotes the Fast Fourier Transform, $B$ denotes the batch size, and $\alpha$ balances the weight between time-domain and frequency-domain precision.

\vspace{-8pt}
\paragraph{Multi-dataset pretraining.}

During large-scale pretraining, Unicorn optimizes Eq.~\eqref{eq:loss} across diverse datasets with varying channel counts.
All model components, including the Spectral Global Guidance module and the Latent Prototype Codebook (with a fixed size $K$), are channel-agnostic.
This ensures that the knowledge encoded in the prototype space is not tied to any specific dataset's channel identities but rather represents universal structural patterns.

\vspace{-8pt}
\paragraph{Regularized finetuning.}
Adaptation to a target domain is achieved by re-aligning local channels to the shared prototype space. To prevent the model from forgetting the universal interaction patterns learned during pretraining, we introduce a structural regularization term. Let $\mathbf{P}$ be the prototype matrix and $\mathbf{P}^{(0)}$ its pretrained state, the finetuning objective is defined as:
\vspace{-4pt}
\begin{equation}
\mathcal{L}=\mathcal{L}_{\text{pred}}+\lambda\left\|\mathbf{P}-\mathbf{P}^{(0)}\right\|_{2}^{2},
\end{equation}
where $\lambda$ governs the trade-off between preservation and adaptation. 
This design encourages the model to adapt primarily via channel-prototype re-assignment rather than re-learning the prototypes.

\vspace{-4pt}
\subsection{Complexity and Scalability}
\label{sec:complexity}
\vspace{-2pt}

The \textit{Channel-Prototype Interaction} module has two main advantages. First, it replaces the dimension-bounded $C \times C$ interactions with a low-rank $C \times K$ interaction pathway.
We here analyze the computational efficiency of Unicorn with respect to the channel dimension $C$, hidden dimension $D$, and the number of prototypes $K$.
Substituting Eq. \eqref{eqn:protype0} into Eq. \eqref{eqn:protype1} yields the following composite form: $\tilde{\mathbf{U}}_{:,m,:} = \mathbf{B}_m \mathbf{A}_m \mathbf{U}_{:,m,:}.$
This formulation reveals that, unlike standard self-attention, which computes a dense $C \times C$ affinity matrix, Unicorn factorizes the inter-channel interaction into a low-rank product $\mathbf{B}_m \mathbf{A}_m$. By routing dependencies through this prototype-mediated bottleneck, the model effectively compresses the attention mechanism:
\begin{itemize}[leftmargin=*]
    \vspace{-4pt}
    \item \textit{Traditional CD models:} Standard Transformer-based CD models rely on exhaustive self-attention across all variables, incurring a quadratic cost of $\mathcal{O}(C^2 \cdot D)$. This cost becomes prohibitive in high-dimensional settings, leading to parameter explosion and severe overfitting when $C > I$.
    \item \textit{Unicorn:} Our mechanism reduces the computational complexity to $\mathcal{O}(C \cdot K \cdot D)$. By mediating dependencies through a fixed-size latent space ($K \ll C$), the model achieves linear scaling with respect to the number of channels. This linear scaling property is particularly valuable for high-dimensional ``wide'' data regimes, where the number of variables significantly outweighs the available temporal observations. 
    \vspace{-4pt}
\end{itemize}

For domain scalability, through the following mechanisms, Unicorn avoids the rigid $C \times C$ parameterization that restricts generalization to fixed dimensionalities, thereby remaining robust to varying channel counts and enabling seamless multi-dataset pretraining across heterogeneous domains:
\begin{itemize}[leftmargin=*]
    \vspace{-4pt}
    \item \textit{Universal interaction vocabulary:} 
    The shared prototype space $\mathbf{P}$ serves as a global codebook, where each prototype encodes abstract, identity-agnostic relational structures that are not tied to the semantics of any single dataset.
    \item \textit{Efficient domain transfer:} 
    The association matrices $\mathbf{A}_m$ and $\mathbf{B}_m$ enable the model to dynamically map heterogeneous channels to these universal prototypes based on their spectral-temporal representations $\mathbf{U}_{:,m,:}$. Adapting to a new domain reduces to learning a lightweight re-alignment of channel-prototype mappings, rather than re-learning correlation structures from scratch.
\end{itemize}
\vspace{-4pt}
\section{Experiments}
\vspace{-2pt}
\subsection{Experimental Setups}
\vspace{-2pt}
\label{sec:experimental_setups}

We evaluate Unicorn through two complementary protocols: (i) \textit{domain-scalable training} across heterogeneous datasets, and (ii) \textit{domain-specific training} for high-dimensional forecasting.

\vspace{-8pt}
\paragraph{Setting I: Cross-domain pretraining.}
This protocol evaluates the model's ability to transfer universal channel relations across heterogeneous datasets. The training follows a two-stage process:
\begin{itemize}[leftmargin=*]
    \vspace{-4pt}
    \item \textit{Pretraining:} 
    Models pretrained on a large-scale multi-market financial dataset comprising $14{,}386$ stocks from the A-share, U.S., and Hong Kong markets.
    \item \textit{Adaptation:}
    The pretrained model is finetuned and evaluated on 
    (i) $587$ A-share stocks (extreme ``wide'' data scenario) and 
    (ii) $88$ NASDAQ-100 constituent stocks.
    \vspace{-4pt}
\end{itemize}
The evaluation period spans from July 1, 2024, to October 20, 2025. 
We use standard metrics in quantitative finance, including the Information Coefficient (IC) and RankIC.
We benchmark against representative CD models (\textit{e.g.}, iTransformer~\citep{itransformer}, TimeBridge~\citep{timebridge}, SOFTS~\citep{softs}) and state-of-the-art CI foundation models (\textit{e.g.}, TimesFM 2.0~\citep{timesfm}, Kronos~\citep{kronos}).

\vspace{-8pt}
\paragraph{Setting II: Domain-specific training.}
This setting evaluates Unicorn within a standard in-domain learning paradigm, where models are trained from scratch on individual benchmarks (Traffic, Crime-Chicago, Electricity, and Wiki-People)
To assess cross-frequency robustness, we also include Traffic-Daily and ECL-Daily, which are generated via temporal aggregation.
The lookback window is fixed at $I=96$, with prediction horizons set to $O=96$ for most datasets and $O=12$ for the high-dimensional Crime-Chicago dataset.

Comprehensive details on data preparation and metrics are provided in Appendices \ref{app:dataset}--\ref{app:metrics}.

\begin{table*}[t]
\centering
\caption{\textbf{Cross-domain scalability results on high-dimensional financial benchmarks.}
We report IC/RankIC ($10^{-2}$, higher is better) over three random seeds.
The symbol $^{\ominus}$ denotes models trained from scratch on the target domain. 
The column $\Delta(10^{-3})$ represents the absolute gain from multi-dataset pretraining, calculated as $(x_{\mathrm{pre}}-x_{\mathrm{scratch}}^{\ominus})\times10^{3}$.
Entries marked ``Not supported'' refer to models whose fixed-dimension designs preclude joint pretraining across heterogeneous domains.
}
\vspace{3pt}
\label{tab:financial_transfer}
\setlength{\tabcolsep}{2pt}
\renewcommand{\arraystretch}{1.08}

\resizebox{\textwidth}{!}{%
\begin{tabular}{l c cc cc cc cc}
\toprule
\multirow{3}{*}{Model} 
& \multirow{3}{*}{Pretrain Data}
& \multicolumn{4}{c}{Performance ($10^{-2}$)}
& \multicolumn{4}{c}{Promotion $\Delta$ $(10^{-3})$} \\
\cmidrule(lr){3-6} \cmidrule(lr){7-10}
& 
& \multicolumn{2}{c}{A-share}
& \multicolumn{2}{c}{NASDAQ-100}
& \multicolumn{2}{c}{A-share}
& \multicolumn{2}{c}{NASDAQ-100} \\
\cmidrule(lr){3-4} \cmidrule(lr){5-6}
\cmidrule(lr){7-8} \cmidrule(lr){9-10}
&  & IC & RankIC & IC & RankIC
& $\Delta$IC & $\Delta$RankIC
& $\Delta$IC & $\Delta$RankIC \\
\midrule

iTransformer & Not supported
& -0.92{\scriptsize$\pm$0.03} & -1.17{\scriptsize$\pm$0.04}
& -0.48{\scriptsize$\pm$0.02} & -0.85{\scriptsize$\pm$0.03}
& -- & -- & -- & -- \\

TimeBridge & Not supported
& 1.83{\scriptsize$\pm$0.04} & 2.31{\scriptsize$\pm$0.03}
& 0.72{\scriptsize$\pm$0.03} & 1.06{\scriptsize$\pm$0.05}
& -- & -- & -- & -- \\

\midrule

PatchTST$^{\ominus}$ & None
& 0.65{\scriptsize$\pm$0.02} & 0.87{\scriptsize$\pm$0.03}
& 0.44{\scriptsize$\pm$0.01} & 0.51{\scriptsize$\pm$0.01}
& -- & -- & -- & -- \\

PatchTST & Financial 
& 0.67{\scriptsize$\pm$0.03} & 0.93{\scriptsize$\pm$0.03}
& 0.39{\scriptsize$\pm$0.01} & 0.44{\scriptsize$\pm$0.02}
& \textcolor{red}{0.2} 
& \textcolor{red}{0.6} 
& \textcolor{green}{-0.5} 
& \textcolor{green}{-0.7} \\

DLinear$^{\ominus}$ & None
& -2.79{\scriptsize$\pm$0.06} & -2.26{\scriptsize$\pm$0.07}
& -3.60{\scriptsize$\pm$0.04} & -3.09{\scriptsize$\pm$0.05}
& -- & -- & -- & -- \\

DLinear & Financial 
& -2.74{\scriptsize$\pm$0.04} & -2.29{\scriptsize$\pm$0.05}
& -3.52{\scriptsize$\pm$0.10} & -2.94{\scriptsize$\pm$0.08}
& \textcolor{red}{0.5} 
& \textcolor{green}{-0.3} 
& \textcolor{red}{0.8} 
& \textcolor{red}{1.5} \\

SOFTS$^{\ominus}$ & None
& -1.58{\scriptsize$\pm$0.03} & -1.39{\scriptsize$\pm$0.02}
& -2.25{\scriptsize$\pm$0.04} & -2.07{\scriptsize$\pm$0.05}
& -- & -- & -- & -- \\

SOFTS & Financial 
& -1.03{\scriptsize$\pm$0.02} & -0.89{\scriptsize$\pm$0.04}
& -1.94{\scriptsize$\pm$0.02} & -1.56{\scriptsize$\pm$0.03}
& \textcolor{red}{5.5} 
& \textcolor{red}{5.0} 
& \textcolor{red}{3.1} 
& \textcolor{red}{5.1} \\

\midrule

TimesFM~2.0 & Time series
& 0.32{\scriptsize$\pm$0.01} & 0.55{\scriptsize$\pm$0.00}
& -0.17{\scriptsize$\pm$0.00} & -0.31{\scriptsize$\pm$0.01}
& -- & -- & -- & -- \\

Kronos & Global financial
& 1.39{\scriptsize$\pm$0.02} & \underline{2.77{\scriptsize$\pm$0.05}}
& 1.22{\scriptsize$\pm$0.04} & 1.89{\scriptsize$\pm$0.02}
& -- & -- & -- & -- \\

\midrule

Unicorn$^{\ominus}$ & None
& \underline{1.85{\scriptsize$\pm$0.03}} & 2.37{\scriptsize$\pm$0.04}
& \underline{1.39{\scriptsize$\pm$0.01}} & \underline{1.95{\scriptsize$\pm$0.04}}
& -- & -- & -- & -- \\

Unicorn & Financial 
& \textbf{2.57{\scriptsize$\pm$0.04}} & \textbf{3.01{\scriptsize$\pm$0.03}}
& \textbf{2.38{\scriptsize$\pm$0.02}} & \textbf{2.51{\scriptsize$\pm$0.02}}
& \textcolor{red}{\textbf{7.2}}
& \textcolor{red}{\textbf{6.4}}
& \textcolor{red}{\textbf{9.9}}
& \textcolor{red}{\textbf{5.6}} \\

\bottomrule
\end{tabular}
}
\vspace{-5pt}
\end{table*}

\begin{table*}[t]
    \centering
    \caption{\textbf{Forecasting performance under domain-specific training.} We report mean MSE and MAE (lower is better) over three random seeds. Despite the significant heterogeneity in application domains and channel semantics across these datasets, our approach consistently achieves state-of-the-art results. Notably, multi-dataset pretraining does not lead to negative transfer.}
    \vspace{3pt}
    \label{tab:benchmark_results}

    \setlength{\tabcolsep}{4pt} 

    \small
    \begin{tabular}{lcccccccccccc}
        \toprule
        \multirow{2}{*}{Model} 
        & \multicolumn{2}{c}{Traffic}
        & \multicolumn{2}{c}{Traffic-Daily}
        & \multicolumn{2}{c}{Crime}
        & \multicolumn{2}{c}{Electricity}
        & \multicolumn{2}{c}{ECL-Daily}
        & \multicolumn{2}{c}{Wiki-People} \\
        \cmidrule(lr){2-3}
        \cmidrule(lr){4-5}
        \cmidrule(lr){6-7}
        \cmidrule(lr){8-9}
        \cmidrule(lr){10-11}
        \cmidrule(lr){12-13}
        & MSE & MAE 
        & MSE & MAE
        & MSE & MAE 
        & MSE & MAE 
        & MSE & MAE
        & MSE & MAE \\
        \midrule
        iTransformer 
            & 0.392 & 0.269 
            & 0.650  & 0.401      
            & 1.202 & 0.667 
            & 0.148 & 0.239 
            & 0.289 & 0.351     
            & 2.449 & 0.893 \\

        PatchTST     
            & 0.360 & 0.249 
            & 0.582 & 0.351      
            & 1.230 & 0.670 
            & 0.132 & 0.227 
            & 0.275 & 0.339      
            & 2.651 & 0.972 \\

        DLinear      
            & 0.410 & 0.282 
            & 0.838 & 0.515     
            & 1.268 & 0.730 
            & 0.141 & 0.236 
            & 0.281 & 0.340   
            & 2.510 & 0.916 \\

        SOFTS 
            & 0.377 & 0.250 
            & 0.623 & 0.390      
            & 1.094 & 0.631 
            & 0.143 & 0.233 
            & 0.290 & 0.355      
            & 2.226 & 0.853 \\

        TimeBridge   
            & \textbf{0.341} & 0.241 
            & 0.514 & \underline{0.300}     
            & 1.116 & 0.624 
            & 0.122 & 0.215 
            & \underline{0.258} & \underline{0.306}     
            & 2.239 & 0.868 \\

        \midrule
        Unicorn$^{\ominus}$      
            & \underline{0.343} & \underline{0.239} 
            & \underline{0.513} & 0.301      
            & \underline{0.925} & \underline{0.527} 
            & \textbf{0.120} & \underline{0.212} 
            & 0.265 & 0.308      
            & \underline{2.107} & \underline{0.831} \\

        Unicorn     
            & \textbf{0.341} & \textbf{0.238} 
            & \textbf{0.497} & \textbf{0.294}      
            & \textbf{0.919} & \textbf{0.526} 
            & \underline{0.121} & \textbf{0.211} 
            & \textbf{0.241} & \textbf{0.295}      
            & \textbf{2.001} & \textbf{0.816} \\

        \bottomrule
    \end{tabular}
    \vspace{-5pt}
\end{table*}

\vspace{-4pt}
\subsection{Results on Cross-Domain Pretraining and Adapation}
\vspace{-2pt}

Table \ref{tab:financial_transfer} presents the cross-market transfer results for high-dimensional financial forecasting, where Unicorn achieves the state-of-the-art performance across both target markets.
Unicorn achieves state-of-the-art results across both target markets, leading to the following key observations.
First, unlike existing \textit{CD architectures} (iTransformer and TimeBridge) that are not applicable for training under shifting channel identities, Unicorn preserves scalability across domains by maintaining parameters independent of $C$.
Second, compared with the \textit{CI foundation models} (TimesFM, Kronos) that are also finetuned on the target benchmarks, Unicorn achieves better performance by flexibly modeling inter-channel correlations.
Finally, compared with other models that support \textit{pretraining on the same datasets}, Unicorn delivers the largest performance improvements (as indicated by $\Delta$), demonstrating its particular strength in learning domain-invariant interaction structures through scalable pretraining.


\vspace{-4pt}
\subsection{Results on Domain-Specific Learning}
\vspace{-2pt}

Table \ref{tab:benchmark_results} summarizes the forecasting performance in a standard supervised regime, where all models are trained from scratch on isolated benchmarks.
Even without the benefit of large-scale pretraining, Unicorn consistently delivers the strongest performance, particularly on high-dimensional and sparse datasets such as Crime-Chicago, Electricity, and Wiki-People.
These results suggest that our channel-prototype interaction acts as a superior structural inductive bias, providing a more robust prior for multivariate dependencies than the identity-bound attention mechanisms of traditional CD models.

We further analyze the transferability of finance-pretrained models to non-financial domains. Given the substantial semantic and distribution shifts between financial markets and physical sensors, the performance gains during zero-shot transfer are understandably modest.
However, it is notable that our multi-dataset pretraining regime does not induce negative transfer, even under extreme domain gaps. This underscores the robustness of the identity-agnostic prototypes, which maintain structural integrity while remaining flexible enough to adapt to disparate data manifold semantics.



\vspace{-4pt}
\subsection{Ablation Studies}
\vspace{-2pt}

We conduct multiple ablation studies to quantify the contribution of each modular component in Unicorn under the domain-specific training regime (Setting II). 
Following the modular architecture presented in Section~\ref{sec:stage2}--\ref{sec:train}, we evaluate three key components of the framework:
\begin{itemize}[leftmargin=*]
    \vspace{-4pt}
    \item \textit{w/o Spectral guidance:} Removes the FAN-based spectral extractor and the aggregate-redistribute pathway. The model relies solely on univariate features in the prototype interaction module.
    \item \textit{w/o Prototype interaction:} Removes the latent prototype codebook and the cascaded attention mechanism, reverting to direct, identity-bound channel-level interaction.
    \item \textit{w/o Spectral loss:} Maintains the full architecture but trains exclusively with a time-domain objective, dropping the frequency-domain term from our hybrid loss function.
    \vspace{-4pt}
\end{itemize}

\begin{figure*}
    \centering
    \includegraphics[width=0.98\linewidth]{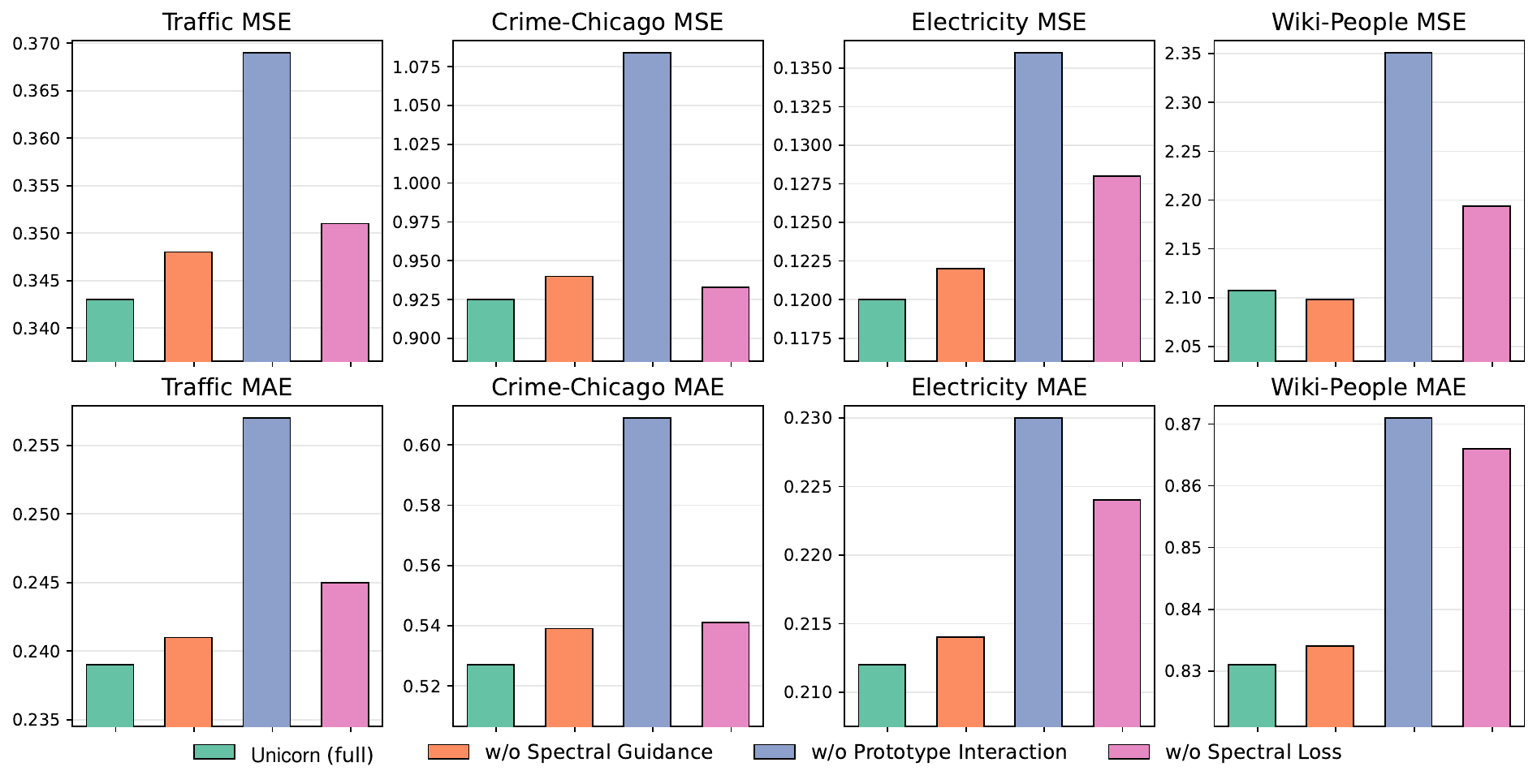}
    \vspace{-8pt}
    \caption{\textbf{Ablation studies under the domain-specific training setting.}
    Removing the Prototype Interaction module leads to the most significant performance drop
    , validating its role for channel correlation modeling. 
    Both Spectral Guidance and the Spectral Loss provide complementary gains.
    }
    \label{fig:ablation}
    \vspace{-5pt}
\end{figure*}

The results, summarized in Figure~\ref{fig:ablation}, validate that all three components are essential for optimal performance.
First, the most significant performance drop occurs when removing \textit{Prototype Interaction}. This suggests that routing dependencies through a shared latent space is far superior to direct channel-level interaction in high-dimensional settings, as it prevents the model from over-fitting to dataset-specific identities.
Besides, removing \textit{Spectral Guidance} leads to consistent degradation. This suggests that frequency-informed global cues provide a stable inter-channel context that helps align channels with related prototypes.
Finally, the omission of \textit{Spectral Loss} harms robustness, particularly in datasets with strong periodicity (\textit{e.g.}, Electricity). This indicates that frequency-domain supervision is vital for preserving the underlying physical structure of the time series.

\vspace{-4pt}
\subsection{Sensitivity to Prototype Size}
\vspace{-2pt}

The number of latent prototypes $K$ controls the capacity of the shared interaction space. To analyze its impact, we evaluate Unicorn with different values of $K$, while keeping all other components and training settings unchanged. 
The results in Table \ref{tab:prototype_size} demonstrate that Unicorn maintains a significant performance edge over the multi-dataset foundation model, Kronos, across all tested values of $K$. However, the optimal value of $K$ is related to the channel scale and market complexity of the target dataset.
For the A-share market ($C=587$), performance metrics (IC and RankIC) improve steadily as $K$ increases. This suggests that larger, more heterogeneous stock pools require a higher-capacity latent space to capture the diverse array of interaction roles present in the market.
Conversely, on the NASDAQ-100 ($C=88$), peak performance is achieved at the smallest prototype size ($K=8$).


\vspace{-4pt}
\subsection{Scalability and Data Efficiency}
\vspace{-2pt}

In Table~\ref{tab:scalability}, we evaluate Unicorn on the A-share dataset (Setting I) under three challenging regimes: channel scalability, robustness to missing data, and few-shot transferrability.

\begin{table}[t]
\centering

\begin{minipage}[t]{0.48\linewidth}
    \centering
    \caption{\textbf{Impact of prototype size $K$ under the cross-domain pretraining setting.}
    Unicorn consistently outperforms Kronos across different prototype sizes.
    The optimal $K$ varies with the channel scale and complexity of the target dataset, suggesting that moderate prototype capacity is important for cross-domain adaptation.
    }
    \vspace{-5pt}
    \label{tab:prototype_size}
    \scriptsize
    \setlength{\tabcolsep}{3pt}
    \renewcommand{\arraystretch}{1.08}
    \resizebox{\linewidth}{!}{
    \begin{tabular}{ccccc}
        \toprule
        \multirow{2}{*}{\# Prototypes} 
        & \multicolumn{2}{c}{A-share} 
        & \multicolumn{2}{c}{NASDAQ-100} \\
        \cmidrule(lr){2-3} \cmidrule(lr){4-5}
        & IC & RankIC & IC & RankIC \\
        \midrule
        Kronos & 0.0139 & 0.0277 & 0.0122 & 0.0189 \\
        \midrule
        2  & 0.0187    & 0.0194 & 0.0185 & 0.0170 \\
        4  & 0.0214    & 0.0236 & 0.0231 & 0.0248 \\
        8  & 0.0249 & 0.0289 & \textbf{0.0241} & \textbf{0.0260} \\
        16 & 0.0257 & 0.0301 & 0.0238 & 0.0251 \\
        32 & \textbf{0.0259} & \textbf{0.0308} & 0.0226 & 0.0247 \\
        \bottomrule
    \end{tabular}
    }
\end{minipage}
\hfill
\begin{minipage}[t]{0.48\linewidth}
    \centering
    \caption{\textbf{Channel scalability, robustness, and few-shot analyses on the A-share dataset.} See text for detailed configurations used in the target-domain finetuning phase and the test phase.}
    \vspace{-5pt}
    \label{tab:scalability}
    \scriptsize
    \setlength{\tabcolsep}{2.5pt}
    \renewcommand{\arraystretch}{0.98}
    \resizebox{\linewidth}{!}{
    \begin{tabular}{lcccc}
        \toprule
        \multirow{2}{*}{Evaluation}
        & \multicolumn{2}{c}{Unicorn}
        & \multicolumn{2}{c}{Kronos} \\
        \cmidrule(lr){2-3} \cmidrule(lr){4-5}
        & IC & RankIC & IC & RankIC \\
        \midrule
        \multicolumn{5}{l}{\textit{Channel scalability: Train = Test}} \\
        25\% channels  & 0.0232 & 0.0274 & 0.0195 & \textbf{0.0243} \\
        50\% channels  & 0.0239 & 0.0292 & \textbf{0.0197} & 0.0236 \\
        100\% channels & \textbf{0.0257} & \textbf{0.0301} & 0.0183 & 0.0231 \\
        \midrule
        \multicolumn{5}{l}{\textit{Channel robustness: Train = full, Test = partial}} \\
        50\% channels  & 0.0244 & 0.0297 & 0.0187 & 0.0229 \\
        25\% channels  & 0.0238 & 0.0280 & 0.0171 & 0.0215 \\
        \midrule
        \multicolumn{5}{l}{\textit{Few-shot finetuning}} \\
        Full FT data   & 0.0257 & 0.0301 & 0.0183 & 0.0231 \\
        50\% FT data   & 0.0241 & 0.0295 & 0.0146 & 0.0185 \\
        25\% FT data   & 0.0238 & 0.0288 & 0.0108 & 0.0160 \\
        \bottomrule
    \end{tabular}
    }
\end{minipage}

\vspace{-6pt}
\end{table}

\vspace{-8pt}
\paragraph{Channel scalability.}
We examine the scalability of Unicorn by varying the number of available variables during both adaptation and inference (we use the same subset of channels in both phases).
Unicorn presents a clear positive scaling effect: as the channel count increases, its performance improves consistently, with IC rising from $0.0232$ (at $25\%$ capacity) to $0.0257$ (full set). In contrast, the CI baseline, Kronos, exhibits performance saturation, with accuracy peaking at $50\%$ and declining thereafter.
This divergence suggests that CI models encounter a ``performance ceiling'' due to their inherent inability to exploit inter-channel dependencies.
Conversely, Unicorn effectively leverages the additional information from larger variable sets to refine its latent interaction features, achieving these gains without the quadratic computational overhead that makes traditional CD models impractical in high-dimensional regimes.

\vspace{-8pt}
\paragraph{Robustness to channel omission.}
We evaluate the robustness of our model to missing channels at inference time, which is a frequent challenge in real-world deployment.
Unicorn demonstrates remarkable stability. Even when $75\%$ of channels are missing at test time, it retains an IC of $0.0238$, which is significantly higher than its CI counterpart.
Interestingly, we observe that the CI model (Kronos), despite its lack of explicit inter-channel modeling, also suffers a noticeable performance decay as channel availability decreases.
This suggests that the performance degradation in both models is not solely a result of losing relational information, but also reflects the stochasticity and variance inherent in the channel sampling process.

\vspace{-8pt}
\paragraph{Few-shot adaptation and data efficiency.}
This final regime evaluates the model's capacity to adapt under severe data scarcity. As the finetuning data is reduced to $25\%$, Unicorn maintains an IC of $0.0238$, representing only a marginal drop from its full-data performance. In contrast, the CI baseline (Kronos) suffers a significant performance collapse, with its IC reducing from $0.0183$ to $0.0108$.
This disparity highlights that, compared to CI models, the latent channel-prototype interaction mechanism serves as a robust structural inductive bias.
Rather than re-learning dependencies from scratch, Unicorn leverages the universal prototypes captured during pretraining. Therefore, few-shot adaptation becomes an efficient task of re-mapping channels to existing prototypes, which requires substantially less target-domain supervision than architectures lacking structural priors.

\vspace{-4pt}
\subsection{Additional Analyses}
\vspace{-2pt}

We provide supplementary analyses in Appendix~\ref{app:additional_analyses}. These include: (i) structural fidelity assessments using Corr-MAE, ACF-MAE, and PSD-L1 to evaluate the preservation of system dynamics (Appendix~\ref{app:structural_validation}); (ii) an investigation into prototype adaptation and drift during finetuning (Appendix~\ref{app:prototype_drift}); (iii) an analysis of cross-domain prototype reuse and functional specialization (Appendix~\ref{app:prototype_semantics}); and (iv) an expanded study on pretraining robustness under distribution shift (Appendix~\ref{app:expanded_pretraining}).

\vspace{-4pt}
\section{Conclusions and Limitations}
\vspace{-2pt}

In this paper, we introduced Unicorn, a unified framework for scalable, multi-dataset pretraining on high-dimensional time series.
By modeling inter-channel correlations within a domain-reusable latent prototype space and leveraging spectral-temporal global guidance for dynamic alignment, UniCorN successfully reconciles the long-standing trade-off between the scalability of channel-independent (CI) architectures and the expressive dependency modeling of channel-dependent (CD) models.
Extensive experiments across financial and physical benchmarks demonstrate that UniCorN consistently achieves superior performance, robustness to missing channels, and strong transferability, suggesting a potential path toward multivariate time series foundation models.

Despite its strengths, several avenues for further development remain: First, this work primarily focuses on offline forecasting with fully observed historical sequences. Future research could investigate Unicorn's performance in real-time streaming environments involving delayed observations or asynchronous updates, which are prevalent in large-scale industrial systems.
In addition, while Unicorn targets continuous-valued time series, extending the framework to mixed-type variables (\textit{e.g.}, categorical or event-based channels) remains unexplored. Addressing these challenges constitutes an important direction for future research.

\section*{Acknowledgments}
This work was supported by the National Natural Science Foundation of China (62250062), the Smart Grid National Science and Technology Major Project (2024ZD0801200), the Shanghai Municipal Science and Technology Major Project (2021SHZDZX0102), the Fundamental Research Funds for the Central Universities, and the Shanghai Jiao Tong University AI for Engineering Initiative (WH410263001/005).

\bibliography{example_paper}
\bibliographystyle{plain}


\appendix

\newpage
\section*{Appendix}

\section{Dataset Details}
\label{app:dataset}

\paragraph{Pretraining dataset.}
The pretraining corpus consists of large-scale financial time series collected from three major equity markets:
the Chinese A-share market, the U.S. stock market, and the Hong Kong stock market.
The dataset spans from January 1, 2010 to December 31, 2023, covering multiple market regimes and economic cycles.
All series are aligned on trading days and contain multi-channel features including price-based indicators
and technical factors.
The total pretraining period contains 3,651 trading days.
The dataset is split chronologically for training and validation to avoid any temporal leakage.

\vspace{-5pt}\paragraph{Evaluation datasets.}
Evaluations are conducted on both financial and public multivariate forecasting benchmarks.
For financial forecasting, we consider two test sets:
(i) an A-share subset containing 587 actively traded stocks,
and (ii) the NASDAQ-100 constituent stocks with 88 equities.
Both test sets use the same evaluation window from July 1, 2024 to October 20, 2025,
covering 341 trading days in total.
Models are trained on historical data before this period and evaluated strictly on the held-out interval. In addition, we evaluate on widely used public multivariate benchmarks,
including Traffic, Electricity (ECL), Crime-Chicago, and Wiki-People.
Traffic and ECL are hourly datasets,
while Crime-Chicago and Wiki-People are monthly and daily datasets, respectively.
Following common practice, all datasets are split chronologically into training,
validation, and test sets.

\vspace{-5pt}\paragraph{Daily-resolution variants.}
To examine robustness under different temporal granularities,
we further construct Traffic-Daily and ECL-Daily by aggregating the original hourly data
into daily frequency.
Specifically, values within each day are aggregated by averaging,
resulting in sequences whose lengths are approximately one twenty-fourth
of the original hourly series while preserving the same channel dimensionality.

\begin{table}[ht]
    \centering
    \caption{\textbf{Summary of datasets used in pretraining and evaluation.}
    ``Dim'' denotes the number of channels.
    ``Length'' refers to the total number of time steps.
    Traffic-Daily and ECL-Daily are obtained by aggregating the original hourly data to daily frequency.}
    \vspace{-3pt}
    \label{tab:dataset_summary}
    \setlength{\tabcolsep}{12pt}
    \begin{tabular}{lcccc}
        \toprule
        Dataset & Domain & Dim & Length & Frequency \\
        \midrule
        A-share (pretrain) & Finance & 4932 & 3,168 & Trading day \\
        U.S. stocks (pretrain) & Finance & 7036 & 3,273 & Trading day \\
        Hong Kong stocks (pretrain) & Finance & 2418 & 3,206 & Trading day \\
        \midrule
        A-share (test) & Finance & 587 & 341 & Trading day \\
        NASDAQ-100 & Finance & 88 & 341 & Trading day \\
        \midrule
        Traffic & Transportation & 862 & 17,545 & Hourly \\
        Traffic-Daily & Transportation & 862 & 729 & Daily \\
        \midrule
        Electricity (ECL) & Energy & 321 & 26,304 & Hourly \\
        ECL-Daily & Energy & 321 & 1,096 & Daily \\
        \midrule
        Crime-Chicago & Public Safety & 1,155 & 260 & Monthly \\
        Wiki-People & Web Traffic & 6,107 & 550 & Daily \\
        \bottomrule
    \end{tabular}
\end{table}

\section{Evaluation Metrics}
\label{app:metrics}

\paragraph{Overview.}
This paper considers two evaluation settings: 
(i) high-dimensional forecasting on public multivariate benchmarks (Setting~II), 
and (ii) cross-market stock return prediction under domain adaptation (Setting~I).
Accordingly, we report error-based forecasting metrics (MSE/MAE) and correlation-based financial metrics (IC/RankIC).

\vspace{-5pt}\paragraph{Notation.}
Let $\mathbf{Y} \in \mathbb{R}^{T \times C}$ denote the ground-truth future sequence
over a forecasting horizon of length $T$ with $C$ channels,
and let $\hat{\mathbf{Y}} \in \mathbb{R}^{T \times C}$ be the corresponding model predictions.
We denote the value at time step $t$ and channel $c$ as $Y_{t,c}$ and $\hat{Y}_{t,c}$, respectively.

\vspace{-5pt}\paragraph{Mean Squared Error (MSE).}
For Setting~II, we evaluate forecasting accuracy using the Mean Squared Error (MSE),
defined as
\begin{equation}
\mathrm{MSE}
=
\frac{1}{TC}
\sum_{t=1}^{T}
\sum_{c=1}^{C}
\left(
Y_{t,c} - \hat{Y}_{t,c}
\right)^2.
\end{equation}
MSE serves as the primary metric for numerical forecasting accuracy, where lower values indicate better performance.

\vspace{-5pt}\paragraph{Mean Absolute Error (MAE).}
We additionally report the Mean Absolute Error (MAE), given by
\begin{equation}
\mathrm{MAE}
=
\frac{1}{TC}
\sum_{t=1}^{T}
\sum_{c=1}^{C}
\left|
Y_{t,c} - \hat{Y}_{t,c}
\right|.
\end{equation}
Compared with MSE, MAE is less sensitive to large deviations and provides a complementary assessment of prediction quality.

\vspace{-5pt}\paragraph{Financial correlation metrics.}
For Setting~I, the objective is to predict cross-sectional stock return signals.
Let $\hat{\mathbf{r}} \in \mathbb{R}^{C}$ denote the predicted scores across $C$ stocks,
and let $\mathbf{r} \in \mathbb{R}^{C}$ be the realized future returns.

\vspace{-5pt}\paragraph{Information Coefficient (IC).}
We measure the linear association between predicted signals and realized returns using the Information Coefficient (IC),
defined as the Pearson correlation:
\begin{equation}
\mathrm{IC}
=
\mathrm{Corr}(\hat{\mathbf{r}}, \mathbf{r})
=
\frac{
\sum_{c=1}^{C}
(\hat{r}_c - \overline{\hat{r}})
(r_c - \bar{r})
}{
\sqrt{
\sum_{c=1}^{C}(\hat{r}_c - \overline{\hat{r}})^2
}
\sqrt{
\sum_{c=1}^{C}(r_c - \bar{r})^2
}
}.
\end{equation}
A higher IC indicates stronger predictive consistency between the model outputs and future returns.

\vspace{-5pt}\paragraph{Rank Information Coefficient (RankIC).}
To further evaluate whether the model preserves the correct cross-sectional ordering,
we report the Rank Information Coefficient (RankIC), computed as the Spearman rank correlation:
\begin{equation}
\mathrm{RankIC}
=
\mathrm{Corr}\big(
\mathrm{rank}(\hat{\mathbf{r}}),
\mathrm{rank}(\mathbf{r})
\big).
\end{equation}
RankIC is particularly important in quantitative finance, as many trading strategies rely on accurate return ranking rather than exact value prediction. We calculate per-day cross-sectional correlation then average over days.

\vspace{-5pt}\paragraph{Metric interpretation.}
All forecasting error metrics (MSE and MAE) are \emph{lower-is-better},
while financial correlation-based metrics (IC and RankIC) are \emph{higher-is-better}.
Together, these metrics provide a rigorous evaluation of both numerical forecasting accuracy
and cross-sectional predictive validity in financial transfer settings.

\section{Baselines}
\label{app:baselines}

To comprehensively evaluate the effectiveness of Unicorn, we compare against a diverse set of representative baselines that span \emph{channel-independent models}, \emph{channel-dependent models}, and \emph{time-series foundation models}. These baselines cover the dominant architectural paradigms in modern multivariate time series forecasting and are evaluated under the same experimental protocols described in the main paper.

\subsection{Channel-Independent Forecasting Models}

Channel-independent (CI) models process each time series independently using shared parameters, without explicitly modeling inter-channel dependencies. This design provides strong scalability and robustness to heterogeneous channel dimensions, making such models particularly competitive in high-dimensional or multi-dataset settings.

\vspace{-5pt}
\paragraph{DLinear~\citep{dlinear}.}
DLinear is a lightweight linear forecasting model that decomposes time series into trend and seasonal components using simple linear projections. Each channel is modeled independently, which leads to excellent computational efficiency and strong robustness in wide-data regimes. However, the absence of explicit inter-channel interaction limits its ability to exploit correlated dynamics across variables.

\vspace{-5pt}
\paragraph{PatchTST~\citep{patchtst}.}
PatchTST adopts a patch-based Transformer architecture where each channel is segmented into temporal patches and processed independently with shared Transformer layers. The model effectively captures long-range temporal dependencies within individual channels while maintaining channel independence. Despite its expressive temporal modeling capacity, PatchTST does not explicitly incorporate inter-channel correlations.

\subsection{Channel-Dependent Forecasting Models}

Channel-dependent (CD) models explicitly model interactions among channels, typically through attention mechanisms or global aggregation modules. While these approaches are expressive, their reliance on identity-specific channel interactions often limits scalability and generalization when channel dimensions vary across datasets.

\vspace{-5pt}\paragraph{iTransformer~\citep{itransformer}.}
iTransformer introduces an inverted attention mechanism that treats channels as tokens and performs attention across the channel dimension. This design enables explicit modeling of inter-channel dependencies and demonstrates strong performance in fixed-channel settings. However, the attention mechanism is tightly coupled to channel identities and dimensionality, making the model unsuitable for scenarios involving heterogeneous channel sets or multi-dataset pretraining.

\vspace{-5pt}\paragraph{TimeBridge~\citep{timebridge}.}
TimeBridge models inter-channel correlations via cointegrated attention over temporally aligned representations. It emphasizes capturing non-stationary and long-range dependencies across channels by learning structured temporal interactions. While effective in domain-specific settings, the architecture assumes a fixed channel set and does not naturally generalize across datasets with varying channel dimensions.

\vspace{-5pt}\paragraph{SOFTS~\citep{softs}.}
SOFTS adopts a series-core fusion strategy, where all channels are aggregated into a global latent core representation and subsequently redistributed back to individual channels. This design explicitly models global inter-channel dependencies while reducing computational overhead compared to full channel-wise attention. Nevertheless, the aggregation process remains tied to the observed channel set, limiting scalability and cross-domain transferability.

\subsection{Time-Series Foundation Models}

Time-series foundation models aim to achieve strong zero-shot or few-shot performance through large-scale pretraining on massive and diverse time series corpora. These models are typically designed with channel-independent architectures to ensure scalability across heterogeneous datasets.

\vspace{-5pt}\paragraph{TimesFM~2.0~\citep{timesfm}.}
TimesFM~2.0 is a decoder-only foundation model pretrained on large-scale time series data. It processes multivariate inputs in a channel-independent manner, sharing parameters across all channels and focusing on learning universal temporal patterns. While it scales effectively to datasets with varying channel dimensions, the model does not explicitly model inter-channel correlations during either pretraining or finetuning.

\vspace{-5pt}\paragraph{Kronos~\citep{kronos}.}
Kronos is a foundation model tailored for financial time series forecasting, pretrained on large-scale global market data. Similar to TimesFM~2.0, it adopts a channel-independent architecture that enables flexible adaptation to different stock universes. Although Kronos exhibits strong generalization and robustness, its CI design limits its ability to capture structured cross-sectional dependencies among financial instruments.



\section{Implementation Details}
\label{app:implementation}


\paragraph{Data preprocessing.}
For all datasets, input time series are normalized using z-score normalization based on statistics computed
from the training split only.
The same normalization parameters are applied to validation and test sets.
For financial datasets, input features include price-based indicators and technical factors,
and labels correspond to next-period returns as defined in Sec.~5.2.
No information from the test period is used during training or validation.

\vspace{-5pt}\paragraph{Training protocol (Setting II).}
For public multivariate forecasting benchmarks (Setting~II),
models are trained from scratch using a sliding-window strategy with a fixed input length $I=96$.
Prediction horizons are set to $O=96$ for Traffic, Electricity, and Wiki-People,
and $O=12$ for Crime-Chicago, consistent with the experimental setup in Sec.~5.1.
The training objective is the spectral-temporal loss in Eq.~(9),
and model selection is performed based on validation MSE.
Final results are reported on the test split using MSE and MAE.

\vspace{-5pt}\paragraph{Pretraining and finetuning protocol (Setting I).}
For cross-market transfer experiments (Setting~I),
Unicorn is pretrained on the multi-market financial corpus described in Appendix~\ref{app:dataset}.
Pretraining optimizes the spectral-temporal objective in Eq.~(9) over heterogeneous channel dimensions.
For downstream adaptation to a target stock pool,
the model is finetuned using the regularized objective in Eq.~(10),
where the pretrained prototype codebook $\mathbf{P}^{(0)}$ is used as the reference.
All reported financial results are evaluated on the held-out test window
using IC and RankIC, as defined in Appendix~\ref{app:metrics}.

\vspace{-5pt}\paragraph{Optimization details.}
All experiments are implemented in PyTorch and trained using the AdamW optimizer.
A cosine learning-rate schedule with linear warm-up is employed.
Early stopping is used based on validation performance,
and the model with the best validation metric is selected for testing.

\vspace{-5pt}\paragraph{Hyperparameter configuration.}
Table~\ref{tab:impl_hparams} summarizes the hyperparameters used in our implementation.
These values are fixed across datasets unless explicitly stated.

\begin{table}[t]
\centering
\caption{\textbf{Hyperparameters}.}
\vspace{-3pt}
\label{tab:impl_hparams}
\setlength{\tabcolsep}{9pt}
\begin{tabular}{lcl}
\toprule
Category & Hyperparameter & Value \\
\midrule
\multirow{4}{*}{Architecture}
& hidden size $D$ & 512 \\
& patch length $S$ & 16 (hourly) / 8 (daily) \\
& prototype size $K$ & 16 \\
& dropout & 0.1 \\
\midrule
\multirow{5}{*}{Training}
& optimizer & AdamW \\
& batch size & 32 \\
& learning rate & $1\times10^{-4}$ \\
& spectral-temporal weight $\alpha$ (Eq.~(9)) & 0.3 \\
& codebook regularization $\lambda$ (Eq.~(10), Setting I) & $1\times10^{-2}$ \\
\bottomrule
\end{tabular}
\end{table}

\section{Additional Analyses}
\label{app:additional_analyses}

We provide additional evidence to further support the structural validity, transfer behavior, and interpretability of Unicorn. Specifically, we include: (i) structure-aware evaluation beyond point-wise forecasting errors, (ii) a quantitative analysis of whether finetuning mainly reuses the pretrained prototype space or rewrites it, (iii) measurements of prototype reuse and semantic coherence across markets, (iv) an expanded pretraining study under domain shift, and (v) qualitative observations of the learned prototypes.

\subsection{Structure-Aware Validation Beyond Point-wise Errors}
\label{app:structural_validation}

While MSE and MAE are standard metrics in multivariate forecasting, they do not directly measure whether a model preserves inter-channel dependency structure, temporal dynamics, or spectral characteristics. To complement the main results, we evaluate Unicorn and five strong baselines on three structure-aware metrics over the non-financial benchmarks.

\vspace{-5pt}\paragraph{Inter-channel correlation preservation.}
We define the correlation error as
\begin{equation}
\mathrm{Corr\text{-}MAE}
=
\frac{1}{C(C-1)}
\sum_{i \neq j}
\left|R_{ij}(Y)-R_{ij}(\hat{Y})\right|,
\end{equation}
where $R(\cdot)$ denotes the channel-wise Pearson correlation matrix.

\vspace{-5pt}\paragraph{Temporal dynamics preservation.}
We measure the discrepancy in autocorrelation structure by
\begin{equation}
\mathrm{ACF\text{-}MAE}
=
\frac{1}{CL}
\sum_{c=1}^{C}
\sum_{\ell=1}^{L}
\left|\rho_c^{(Y)}(\ell)-\rho_c^{(\hat{Y})}(\ell)\right|,
\end{equation}
where $\rho_c(\ell)$ is the autocorrelation at lag $\ell$.

\vspace{-5pt}\paragraph{Spectral consistency.}
We further quantify the discrepancy in normalized power spectra:
\begin{equation}
\mathrm{PSD\text{-}L1}
=
\frac{1}{C}
\sum_{c=1}^{C}
\left\|\tilde{S}_c(Y)-\tilde{S}_c(\hat{Y})\right\|_1,
\end{equation}
where $\tilde{S}_c(\cdot)$ denotes the normalized power spectrum of channel $c$.

Table~\ref{tab:structural_validation} shows that Unicorn consistently achieves the best performance across all three structure-aware metrics on Traffic, Crime-Chicago, and Electricity. This indicates that the gain of Unicorn is not limited to lower point-wise forecasting error; it also better preserves the underlying inter-channel dependency structure, temporal evolution patterns, and spectral behavior of multivariate time series.

\begin{table*}[t]
    \centering
    \caption{\textbf{Structure-aware validation on non-financial benchmarks.} We report Corr-MAE, ACF-MAE, and PSD-L1, where lower is better. Unicorn consistently outperforms all baselines in preserving inter-channel correlations, temporal dynamics, and spectral characteristics.}
    \vspace{3pt}
    \label{tab:structural_validation}
    \setlength{\tabcolsep}{1.15pt}
    \small
    \begin{tabular}{lccccccccc}
        \toprule
        \multirow{2}{*}{Model}
        & \multicolumn{3}{c}{Traffic}
        & \multicolumn{3}{c}{Crime-Chicago}
        & \multicolumn{3}{c}{Electricity} \\
        \cmidrule(lr){2-4} \cmidrule(lr){5-7} \cmidrule(lr){8-10}
        & Corr-MAE & ACF-MAE & PSD-L1
        & Corr-MAE & ACF-MAE & PSD-L1
        & Corr-MAE & ACF-MAE & PSD-L1 \\
        \midrule
        iTransformer & 0.126 & 0.102 & 0.159 & 0.177 & 0.128 & 0.198 & 0.090 & 0.081 & 0.124 \\
        PatchTST     & 0.121 & 0.098 & 0.154 & 0.181 & 0.132 & 0.203 & 0.087 & 0.079 & 0.120 \\
        DLinear      & 0.139 & 0.114 & 0.173 & 0.190 & 0.141 & 0.216 & 0.095 & 0.085 & 0.129 \\
        SOFTS        & 0.112 & 0.094 & 0.147 & 0.152 & 0.113 & 0.178 & 0.086 & 0.078 & 0.118 \\
        TimeBridge   & 0.108 & 0.091 & 0.140 & 0.161 & 0.119 & 0.184 & 0.081 & 0.074 & 0.111 \\
        Unicorn      & \textbf{0.101} & \textbf{0.087} & \textbf{0.133} & \textbf{0.134} & \textbf{0.101} & \textbf{0.161} & \textbf{0.074} & \textbf{0.069} & \textbf{0.103} \\
        \bottomrule
    \end{tabular}
\end{table*}

\subsection{Prototype Drift and Reassignment During Finetuning}
\label{app:prototype_drift}

A central claim of Unicorn is that target-domain finetuning mainly adapts the \emph{channel-to-prototype alignment}, rather than relearning the shared prototype codebook from scratch. To verify this hypothesis, we quantify both \emph{prototype drift} and \emph{reassignment drift} after adaptation.

Recall that in Section~3.4, for each sample $x$ and temporal index $m$, Unicorn performs prototype-mediated interaction through a prototype-to-channel redistribution stage. Specifically, given the augmented channel representations $U_{:,m,:}^{(x,d)}$ and the prototype summaries $Z_m^{(x,d)}$, the redistribution matrix is defined as
\begin{equation}
B_m^{(x,d)}
=
\operatorname{softmax}
\!\left(
\frac{
U_{:,m,:}^{(x,d)}
\left(Z_m^{(x,d)}\right)^\top
}{
\sqrt{2D}
}
\right)
\in \mathbb{R}^{C_d \times K},
\end{equation}
where $d$ denotes the target domain, $C_d$ is the number of channels in that domain, and $K$ is the number of latent prototypes. Each row of $B_m^{(x,d)}$ represents how one target-domain channel is redistributed over the shared prototype basis. Therefore, $B_m^{(x,d)}$ directly characterizes the channel-to-prototype alignment induced by the current model.

Let $P^{(0)} \in \mathbb{R}^{K \times 2D}$ denote the pretrained prototype codebook before target-domain adaptation, and let $P^{(d)}$ denote the prototype codebook after finetuning on domain $d$. We define the \emph{prototype drift} as
\begin{equation}
\delta_P^{(d)}
=
\frac{\left\|P^{(d)} - P^{(0)}\right\|_F}{\left\|P^{(0)}\right\|_F}.
\end{equation}
This quantity measures how much the shared prototype basis itself changes during finetuning.

To quantify how much adaptation happens through channel reassignment rather than codebook rewriting, we further define the \emph{reassignment drift}. Let $S_d$ be the evaluation set of target domain $d$, and let $M$ be the number of long-term temporal tokens used in Section~3.2. For each sample $x \in S_d$ and token index $m$, we compare the redistribution matrix before and after finetuning:
\begin{itemize}[leftmargin=*]
    \item $B_m^{(x,d,0)}$: computed by the pretrained model \emph{before} target-domain finetuning;
    \item $B_m^{(x,d,\mathrm{ft})}$: computed by the adapted model \emph{after} finetuning on domain $d$.
\end{itemize}
We then define
\begin{equation}
\delta_B^{(d)}
=
\frac{1}{|S_d|M}
\sum_{x \in S_d}
\sum_{m=1}^{M}
\frac{
\left\|
B_m^{(x,d,\mathrm{ft})}
-
B_m^{(x,d,0)}
\right\|_1
}{
C_d
}.
\end{equation}
Here, the $\ell_1$ distance is normalized by $C_d$ so that the magnitude of $\delta_B^{(d)}$ remains comparable across domains with different channel dimensions. Intuitively, $\delta_B^{(d)}$ measures how much the target-domain channels are re-aligned to the shared prototype space during adaptation.

In addition to standard finetuning (\textit{Full FT}), we consider two controlled variants:
\begin{itemize}[leftmargin=*]
    \item \textbf{Frozen $P$}: keep the pretrained prototype codebook fixed during finetuning;
    \item \textbf{Reinit $P$}: randomly reinitialize the prototype codebook before finetuning.
\end{itemize}

Table~\ref{tab:prototype_drift} supports the intended transfer mechanism of Unicorn. First, \textit{Frozen $P$} preserves most of the gain, indicating that the pretrained prototype space already captures reusable interaction structure. Second, \textit{Reinit $P$} leads to a clear degradation, suggesting that the learned prototype basis cannot be replaced by an arbitrary initialization. Third, across both A-share and NASDAQ-100, we observe $\delta_B^{(d)} \gg \delta_P^{(d)}$, which indicates that target-domain adaptation primarily happens through reassignment over the shared prototype basis, rather than by substantially rewriting the prototype codebook itself.
\begin{table}[t]
    \centering
    \caption{\textbf{Prototype drift and reassignment drift under domain adaptation.} Full FT denotes standard finetuning. Frozen $P$ fixes the pretrained prototype codebook during finetuning, and Reinit $P$ randomly reinitializes it before adaptation. Results show that transfer mainly happens through reassignment rather than rewriting the shared prototype space.}
    \vspace{-3pt}
    \label{tab:prototype_drift}
    \setlength{\tabcolsep}{9pt}
    \begin{tabular}{lccccc}
        \toprule
        Target
        & Full FT
        & Frozen $P$
        & Reinit $P$
        & $\delta_P^{(d)}$
        & $\delta_B^{(d)}$ \\
        \midrule
        A-share     & 0.0257 & 0.0251 & 0.0203 & 0.07 & 0.41 \\
        NASDAQ-100  & 0.0238 & 0.0232 & 0.0181 & 0.09 & 0.47 \\
        \bottomrule
    \end{tabular}
\end{table}

\subsection{Cross-domain Reuse and Semantic Coherence of Prototypes}
\label{app:prototype_semantics}

To further examine whether the learned prototypes exhibit reusable and non-trivial structure, we quantify both cross-domain reuse and within-prototype statistical coherence.

For each channel $c$ in domain $d$, we first compute its average prototype-assignment profile:
\begin{equation}
\bar{b}_c^{(d)}
=
\frac{1}{|S_d|M}
\sum_{x \in S_d}
\sum_{m=1}^{M}
B_m^{(x)}(c,:)
\in \mathbb{R}^{K},
\end{equation}
and define its dominant prototype by
\begin{equation}
k_c^{(d)} = \arg\max_k \bar{b}_c^{(d)}(k).
\end{equation}

We summarize domain-level prototype usage by
\begin{equation}
q_d(k)
=
\frac{1}{|S_d|MC_d}
\sum_{x \in S_d}
\sum_{m=1}^{M}
\sum_{c=1}^{C_d}
B_m^{(x)}(c,k),
\end{equation}
and define the cross-domain usage similarity as
\begin{equation}
\mathrm{Cos}(d,d')
=
\frac{q_d^\top q_{d'}}{\|q_d\|_2 \|q_{d'}\|_2}.
\end{equation}

To test whether channels grouped under the same prototype are statistically coherent, let $s_c$ be a raw-series statistic such as dominant period, volatility, or lag-1 autocorrelation. We define
\begin{equation}
\rho_s
=
\frac{
\frac{1}{K}
\sum_{k=1}^{K}
\mathrm{Var}(s_c \mid k_c^{(d)} = k)
}{
\mathrm{Var}_{\mathrm{rand}}(s_c)
},
\end{equation}
where $\mathrm{Var}_{\mathrm{rand}}(s_c)$ is the matched random-partition baseline.

Finally, to test whether prototypes collapse into domain labels, we define
\begin{equation}
\mathrm{Purity}
=
\frac{1}{K}
\sum_{k=1}^{K}
\max_d p(d \mid k),
\end{equation}
where $p(d \mid k)$ is the fraction of channels assigned to prototype $k$ from domain $d$.

Table~\ref{tab:prototype_semantics} suggests that the learned prototype space is both reusable and non-degenerate. The high cosine similarity between A-share and NASDAQ-100 indicates that the two markets rely on broadly similar prototype usage patterns. At the same time, the moderate purity score shows that prototypes do not simply collapse into market-specific clusters. Moreover, all $\rho_s < 1$ indicate that channels grouped under the same prototype are more statistically coherent than random partitions.

\begin{table}[t]
    \centering
    \caption{\textbf{Cross-domain reuse and semantic coherence of learned prototypes.} Higher $\mathrm{Cos}$ indicates more similar prototype usage across domains. Lower $\rho_s$ indicates stronger within-prototype coherence relative to a matched random baseline.}
    \vspace{-3pt}
    \label{tab:prototype_semantics}
    \setlength{\tabcolsep}{20pt}
    \begin{tabular}{lc}
        \toprule
        Metric & Value \\
        \midrule
        $\mathrm{Cos}(\text{A-share}, \text{NASDAQ-100})$ & 0.87 \\
        $\mathrm{Purity}$ & 0.58 \\
        $\rho_{\text{period}}$ & 0.68 \\
        $\rho_{\text{vol}}$ & 0.74 \\
        $\rho_{\text{acf1}}$ & 0.79 \\
        \bottomrule
    \end{tabular}
\end{table}

\subsection{Expanded Pretraining Study Under Domain Shift}
\label{app:expanded_pretraining}

In the main paper, we observe that financial-only pretraining yields limited gains on non-financial benchmarks such as Traffic and Electricity. To better understand this phenomenon, we conduct an additional experiment that expands the pretraining corpus by including the training split of the target dataset itself.

We compare three settings:
\begin{itemize}[leftmargin=*]
    \item \textbf{None}: train from scratch on the target dataset;
    \item \textbf{Financial}: pretrain on the financial corpus and then finetune on the target dataset;
    \item \textbf{Financial + Target}: jointly pretrain on the financial corpus and the training split of the target dataset before finetuning.
\end{itemize}

The results in Table~\ref{tab:expanded_pretraining} reveal two important trends. First, financial-only pretraining does not degrade performance relative to training from scratch, suggesting that the prototype-mediated pathway is robust to mismatched source distributions and does not induce negative transfer. Second, when pretraining covers the target domain, Unicorn benefits more clearly, indicating that transfer effectiveness depends on how well the pretraining distribution matches the target interaction patterns.

\begin{table}[t]
    \centering
    \caption{\textbf{Expanded pretraining under domain shift.} Financial-only pretraining yields limited but non-destructive transfer to physical domains, while joint pretraining with target-domain coverage leads to clearer improvements.}
    \vspace{-3pt}
    \label{tab:expanded_pretraining}
    \setlength{\tabcolsep}{15pt}
    \begin{tabular}{lccc}
        \toprule
        Pretrain & Finetune & MSE & MAE \\
        \midrule
        None & Traffic & 0.343 & 0.239 \\
        Financial & Traffic & 0.341 & 0.238 \\
        Financial + Traffic & Traffic & \textbf{0.335} & \textbf{0.234} \\
        \midrule
        None & Electricity & 0.120 & 0.212 \\
        Financial & Electricity & 0.121 & 0.211 \\
        Financial + Electricity & Electricity & \textbf{0.119} & \textbf{0.205} \\
        \bottomrule
    \end{tabular}
\end{table}


\end{document}